\pdfoutput=1

\documentclass[11pt]{article}

\usepackage[final]{acl}

\usepackage{times}
\usepackage{latexsym}

\usepackage[T1]{fontenc}

\usepackage[utf8]{inputenc}

\usepackage{microtype}

\usepackage{inconsolata}

\usepackage{graphicx}

\usepackage{subfigure}
\usepackage{booktabs} 

\usepackage[ruled,vlined]{algorithm2e}

\usepackage{amsmath}
\usepackage{amsthm}
\usepackage{dsfont}
\usepackage{amssymb}
\usepackage{amscd}
\usepackage{mathtools}
\mathtoolsset{showonlyrefs=true}
\usepackage{enumerate}
\usepackage{ bbold }

\usepackage{xcolor}

\usepackage{tikz}
\usepackage{tcolorbox}
\usepackage{lipsum}

\newlength{\RoundedBoxWidth}
\newsavebox{\GrayRoundedBox}
\newenvironment{GrayBox}[1][\dimexpr\columnwidth-4.5ex]%
   {\setlength{\RoundedBoxWidth}{\dimexpr#1}
    \begin{lrbox}{\GrayRoundedBox}
       \begin{minipage}{\RoundedBoxWidth}}%
   {   \end{minipage}
    \end{lrbox}
    \begin{center}
    \begin{tikzpicture}%
       \draw node[draw=black,fill=black!10,rounded corners,%
             inner sep=2ex,text width=\RoundedBoxWidth]%
             {\usebox{\GrayRoundedBox}};
    \end{tikzpicture}
    \end{center}}

\newtheorem{theorem}{Theorem}

\newtheorem{prop}{Property}

\newcommand{\kl}[2]{\operatorname*{KL}(#1||#2)}

\newcommand{\R}{\mathbb{R}}

\newcommand{\E}{\mathbb{E}}

\newcommand{\piref}{\pi_\text{ref}}

\newcommand{\copg}{\texttt{CoPG}}

\usepackage{todonotes}
\usepackage[normalem]{ulem}

\title{Contrastive Policy Gradient:\\ Aligning LLMs on sequence-level scores in a supervised-friendly fashion}

\author{
 \textbf{Yannis Flet-Berliac$^\dagger$},
 \textbf{Nathan Grinsztajn$^\dagger$},
 \textbf{Florian Strub},
 \textbf{Bill Wu},
 \textbf{Eugene Choi},
 \\
 \textbf{Chris Cremer},
 \textbf{Arash Ahmadian},
 \textbf{Yash Chandak },
 \textbf{Mohammad Gheshlaghi Azar},
 \\
 \textbf{Olivier Pietquin},
 \textbf{Matthieu Geist$^*$}
\\
\\
 Cohere
}

\begin{document}
\maketitle
\begin{abstract}
Reinforcement Learning (RL) has been used to finetune Large Language Models (LLMs) using a reward model trained from preference data, to better align with human judgment. The recently introduced direct alignment methods, which are often simpler, more stable, and computationally lighter, can more directly achieve this. However, these approaches cannot optimize arbitrary rewards, and the preference-based ones are not the only rewards of interest for LLMs (\textit{e.g.}, unit tests for code generation or textual entailment for summarization, among others). RL-finetuning is usually done with a variation of policy gradient, which calls for on-policy or near-on-policy samples, requiring costly generations. We introduce \emph{Contrastive Policy Gradient}, or \copg, a simple and mathematically principled new RL algorithm that can estimate the optimal policy even from off-policy data. It can be seen as an off-policy policy gradient approach that does not rely on important sampling techniques and highlights the importance of using (the right) state baseline. We show this approach to generalize the direct alignment method IPO (identity preference optimization) and classic policy gradient. We experiment with the proposed \copg{}  on a toy bandit problem to illustrate its properties, as well as for finetuning LLMs on a summarization task, using a learned reward function considered as ground truth for the purpose of the experiments.
\looseness=-1
\end{abstract}

\def\thefootnote{$\dagger$}\footnotetext{Equal contribution.}
\def\thefootnote{$\star$}\footnotetext{Corresponding author: \texttt{matthieu@cohere.com}.}
\def\thefootnote{\arabic{footnote}}

\section{Introduction}

Reinforcement Learning from Human Feedback (RLHF) \citep{christiano2017deep} is a classic finetuning step intended at aligning a Large Language Model (LLM) with human judgment \citep{ouyang2022training}. The underlying principle is to learn a reward model from a preference dataset, and to optimize this reward with a regularized Reinforcement Learning (RL) approach \citep{fox2015taming,jaques2017sequence,geist2019theory}, usually a Policy Gradient (PG) approach \citep{williams1991function} or a variation like Proximal Policy Optimization (PPO) \citep{schulman2017proximal}. These methods require on-policy or near-on-policy samples, and thus require costly generations from the LLM. They can also be hard to tune and computationally heavy, for example through the use of an additional value network.
\looseness=-1

More recently, the field of direct alignment methods has surged with the introduction of Direct Preference Optimization (DPO) \citep{rafailov2023direct}, Sequence Likelihood Calibration (SLiC-HF) \citep{zhao2023slic} or Identity Preference Optimization (IPO) \citep{azar2024general}. 
These approaches allow directly learning a policy optimizing for preferences, from a given preference dataset, in an offline manner and without using a proxy reward function. They are usually considered simpler, more stable, and computationally more lightweight than classic RLHF. However, by design, they cannot optimize for arbitrary reward functions.

We posit that preference-based rewards are not the only rewards worth considering when finetuning an LLM. Not everything can be measured through preferences, which are also costly to label. 
Such examples are using unit tests as a reward for code generation \citep{le2022coderl} or a reward measuring textual-entailment for summarization \citep{roit-etal-2023-factually}. The aim of this paper is to propose an RL approach able to optimize an arbitrary reward while being as simple as direct alignment.
It is important to note that we do not introduce any specific reward here. Our intent is to provide a convenient and efficient tool for optimizing an arbitrary reward function. In particular, we take inspiration from the contrastive learning objective, which had tremendous success in the self-supervised learning technique~\citep {oord2018representation,chen2020simple}, and extend it to RL techniques.
\looseness=-1

To this end, we introduce \emph{Contrastive Policy Gradient}, or \copg. It minimizes a supervised-friendly loss, of which we show the optimal policy of interest (optimizing the initial RL problem) to be the unique minimizer. It can be interpreted as a form of off-policy policy gradient, not relying on importance sampling (an approach that can easily lead the gradient variance to explode), but exploiting a specific state baseline that can be seen as a contrastive term to the reward being optimized. 
Our proposed approach is versatile as it regroups IPO and policy gradient as special cases. Notably, we obtain as a special case an offline off-policy generalization of Reinforce Leave-One-Out (RLOO) \citep{kool2019buy,ahmadian2024back}.
To illustrate its properties, we experiment with the proposed \copg{} on a toy bandit problem. We also test it for finetuning an LLM on a summarization task. For this case, we train a reward model from the preference dataset and consider it the ground truth to be optimized.

\section{Background}

We denote a prompt $x$ and a generation $y$, and we call the LLM to be trained a policy $\pi(y|x)$. We assume the prompts to be sampled according to some unknown distribution $\rho$. We also assume to have access to some reference model $\piref$, typically the LLM pretrained and supervised-finetuned (SFT model), used both for initializing $\pi$ and regularizing the RL problem. We consider having access to a reward model $R(x,y)$ to be maximized under $\pi$, with some regularization toward the reference model through a KL-divergence $\kl{\pi(\cdot|x)}{\piref(\cdot|x)}$. To lighten notations, we drop the prompt $x$ in the main text (it appears explicitly again in the proofs in Appx.~\ref{appx:proofs}). 

The regularized RL problem consists in maximizing $J(\pi) = \E_{y\sim\pi}[R(y)] - \beta\kl{\pi}{\piref}$. Let's write the regularized reward
\looseness=-1
\begin{equation}
    R_\beta^\pi(y) = R(y) - \beta\ln\frac{\pi(y)}{\piref(y)}, %
    \label{eq:reg_reward}
\end{equation}
the RL problem can equivalently be written as
\begin{equation}
    J(\pi) = \E_{y\sim\pi}[R_\beta^\pi(y)].
    \label{eq:pg_objevctive}
\end{equation}
A classic approach is policy gradient, which maximizes $J$ by gradient ascent. This is not a supervised-friendly loss because the expectation depends on the optimized policy, not on some fixed dataset of generations. 
The gradient is given by
\begin{equation}
    \nabla J(\pi) = \E_{y\sim\pi}[R_\beta^\pi(y) \nabla\ln\pi(y)].
\end{equation}
In practice, an empirical estimate of this gradient requires fresh generations from $\pi$, making it a costly method. It is common to subtract a baseline $\color{blue} b$:
\begin{equation}
    \nabla J(\pi) = \E_{y\sim\pi}[(R_\beta^\pi(y) {\color{blue} - b} )\nabla\ln\pi(y)].
    \label{eq:gradient_pg}
\end{equation}
This does not bias the gradient as long as it does not depend on $y$ (because $\E_{y\sim\pi}[\nabla\ln\pi(y)]=0$), and this is generally introduced to reduce the variance of the empirical gradient \citep{greensmith2001variance}. A classic baseline is an estimate of the expected reward (that is the value, $b \approx \E_{y\sim\pi}[R_\beta^\pi(y)]$).

Objective~\eqref{eq:pg_objevctive} can be made supervised-friendly by relying on some fixed sampling distribution $\mu$ (\textit{e.g.}, underlying a dataset), by using importance sampling. Indeed, we have
\looseness=-1
\begin{equation}
    J(\pi) = \E_{y\sim{\color{blue}\mu}}[{\color{blue}\frac{\pi(y)}{\mu(y)}}R_\beta^\pi(y)].
    \label{eq:pg_is_objective}
\end{equation}
The related gradient is, with a baseline here,
\begin{equation}
    \nabla J(\pi) = \E_{y\sim{\color{blue}\mu}}[{\color{blue}\frac{\pi(y)}{\mu(y)}}(R_\beta^\pi(y) {\color{blue} - b} )\nabla\ln\pi(y)].
\end{equation}
The corresponding empirical gradient suffers from a larger variance: when the policy $\pi$ becomes different from the sampling model $\mu$ (which happens, as $\pi$ is trained), the probability ratio can explode. Training is thus not stable or efficient. Moreover, this requires having access to the probabilities $\mu(y)$, which may not be possible,  for example when the underlying generations have been made by humans rather than by an LLM.
\looseness=-1

An approach for alleviating this stability issue is to clip the probability ratio in objective \eqref{eq:pg_is_objective}, to kill the gradient whenever $\pi$ becomes too different from $\mu$. This is the core idea behind PPO, which considers for $\mu$ older copies of the $\pi$ network and uses a value estimate as the baseline. However, this still requires fresh generations, even if they can be used for more than one gradient step. %

\section{Contrastive Policy Gradient}

\subsection{General objective}

Contrastive approach requires pairs of generations, which do not need to be ranked as in RLHF. For a pair of independent generations $(y,y')$, we introduce the following sample loss:
\looseness=-1
\begin{GrayBox}
\begin{align}
        \ell_\copg&(y,y';\pi) = \label{eq:loss_copg}
        \\
        & \quad\left(R_{\beta/2}^\pi(y) - R_{\beta/2}^\pi(y')\right) \ln \frac{\pi(y)}{\piref(y)}
        \\
        &+ \left(R_{\beta/2}^\pi(y') - R_{\beta/2}^\pi(y)\right) \ln \frac{\pi(y')}{\piref(y')}.
\end{align}
\end{GrayBox}
This can be seen as a weighted log-likelihood, where the weight is the reward of the generation contrasted with the reward of an independent generation, and it's symmetric.

Let $\mu_1$ and $\mu_2$ be some independent distributions (for example underlying a dataset of pairs of generations), that do not need to be known analytically (contrary to policy gradient with importance sampling), and that can be the same too. The objective to be maximized is then 
\begin{equation}
    L(\pi) = \E_{y\sim\mu_1, y'\sim\mu_2}[\ell_\copg(y,y';\pi)].
    \label{eq:copg_objective_function}
\end{equation}
To get more insights, let's rewrite this objective. First, write the expected reward over $\mu$ as
\begin{equation}
    \overline{R_{\beta/2}^\pi}^{\mu} = \E_{y\sim\mu}[R_{\beta/2}^\pi(y)].
\end{equation}
Notice that in RL terms, this is not strictly speaking the value, as the expectation is under $\mu$ rather than $\pi$.
Then, $L$ can be rewritten as
\begin{align}
    L(\pi) &= \E_{y~\sim\mu_1}[\left(R_{\beta/2}^\pi(y) - \overline{R_{\beta/2}^\pi}^{\mu_2}\right) \ln \frac{\pi(y)}{\piref(y)}]
    \\
    &+ \E_{y'~\sim\mu_2}[\left(R_{\beta/2}^\pi(y') - \overline{R_{\beta/2}^\pi}^{\mu_1}\right) \ln \frac{\pi(y')}{\piref(y')}].
\end{align}
Again, this can be seen as a weighted log-likelihood, where the reward weighting the log-likelihood under one distribution is contrasted with the expected reward under the other distribution. This loss is supervised-friendly, as it does not involve sampling from the trained policy.
\looseness=-1

The natural question is whether maximizing this objective $L$ solves the intended problem~\eqref{eq:pg_objevctive}, and thus maximizes any language scores. The answer is positive, and we deferred all proofs to Appx.~\ref{appx:proofs}:
\begin{theorem}[\copg{} solves the right problem]
    \label{thm:right_solution}
    Assume that $\piref$, $\mu_1$ and $\mu_2$ all have the same support. Then, the unique maximizer of $L(\pi)$, defined Eq.~\eqref{eq:copg_objective_function}, is
    $\pi_*(y)\propto \piref(y)\exp\frac{R(y)}{\beta}$,
    which is also the unique maximizer of $J(\pi)$.
    \looseness=-1
\end{theorem}

To shed more light on the relationship to policy gradient, let's consider the gradient of $L(\pi)$. By simple calculus (taking care of the fact that $R^\pi_{\beta/2}$ does depend on $\pi$), one obtains:
\begin{align}
    \nabla &L(\pi) = \E_{y~\sim\mu_1}[\left(R_{\beta}^\pi(y) - \overline{R_{\beta}^\pi}^{\mu_2}\right) \nabla\ln \pi(y)]
    \\
    &+ \E_{y'~\sim\mu_2}[\left(R_{\beta}^\pi(y') - \overline{R_{\beta}^\pi}^{\mu_1}\right) \nabla \ln \pi(y')].
    \label{eq:copg_gradient_expected}
\end{align}
When compared to Eq.~\eqref{eq:gradient_pg}, the classic policy gradient with baseline, we obtain a sum of two policy-like gradients, however with striking differences. First, the expectation is not according to the learnt policy $\pi$, but according to either $\mu_1$ or $\mu_2$, meaning that it can be understood as a sound off-policy policy gradient. Second, there is a baseline, the contrastive term, which is the expected reward but according to the other distribution (which can be the same if both are identically distributed). Crucially, it cannot be any baseline (because $\E_{y\sim\mu}[\nabla\ln\pi(y)]\neq 0$ in general), it must be this specific one.
\looseness=-1

Overall, the proposed objective function \eqref{eq:copg_objective_function}, alongside with the strong result of Thm.~\ref{thm:right_solution}, thanks to the specific form of the gradient~\eqref{eq:copg_gradient_expected}, tells us that policy gradient can be safely applied to off-policy data, without the introduction of a correcting importance sampling term, if we use the right baseline, that is the contrastive term depicted above. The relationship to policy gradient can be made even clearer for the specific case $\mu_1=\mu_2$, to be compared again to Eq.~\eqref{eq:gradient_pg}:
\begin{align}
    \nabla L(\pi) \stackrel{\mu_1=\mu_2}{=} 2 \E_{y\sim\mu}[\left(R_{\beta}^\pi(y) - \overline{R_{\beta}^\pi}^{\mu}\right) \nabla\ln \pi(y)].
\end{align}

\subsection{A simple sample-based objective}
\label{sec:sample_based_objective}

To obtain a practical algorithm, one has to choose what $\mu_1$ and $\mu_2$ are, if they are different or the same, and how to estimate the expectations.
The outer expectations can be estimated using Monte Carlo (forming a batch for each gradient step). Depending on the nature of $\mu_i$, the inner expectations underlying the terms $\overline{R_{\beta}^\pi}^{\mu_i}$ can be estimated using a single (or multi-) sample Monte Carlo estimate, or even possibly by learning an associated value network. 
Regarding the distributions $\mu_1$ or $\mu_2$, the main constraint is that they share the same prompts. It can be a dataset, generations from the current policy, generations from another policy, or coming from a replay buffer as commonly done for off-policy RL methods \citep{mnih2015human}. One could also choose a hybrid approach, where $\mu_1$ is for example a dataset of good but suboptimal generations while $\mu_2$ comes from a replay buffer collecting past generations of the trained policy. This is reminiscent of RL from demonstrations, which has been shown to be beneficial in the classic RL setting \citep{piot2014boosted,hester2018deep}.
\looseness=-1

All these choices may impact the stability and the efficiency of the resulting algorithm. We leave these interesting research directions for future works and focus here on the simple case where we learn in an offline manner from a given  dataset, reminiscent of the now commonly used direct alignment methods, except that we do not need rankings. 

\begin{GrayBox}
Let $\mathcal{D} = \{(y_j,y'_j)_{1\leq j\leq n}\}$ be a dataset of pairs of scored generations with identical prompts. \copg{} minimizes the following empirical loss:
\begin{equation}
    \hat{L}(\pi) = \frac{1}{n} \sum_{j=1}^n \ell_\copg(y_j,y_j';\pi).
    \label{eq:practical_copg_objective}
\end{equation}
with $\ell_\copg$ being defined in Eq.~\eqref{eq:loss_copg}. 
\end{GrayBox}
It is a simple supervised-friendly objective function that can be minimized by performing gradient ascent on mini-batches sampled from the dataset. The gradient can readily be obtained by auto-differentiation (contrary to the gradient of Eq.~\eqref{eq:pg_objevctive}, due to the dependency of the expectation to the optimized policy), but we give it for a pair of generations for completeness:
\begin{align}
        \nabla \ell_\copg&(y,y';\pi) = \left(R_{\beta}^\pi(y) - R_{\beta}^\pi(y')\right) \nabla \ln \pi(y)
        \\
        &+ \left(R_{\beta}^\pi(y') - R_{\beta}^\pi(y)\right) \nabla \ln \pi(y').
        \label{eq:grad_sample_copg}
\end{align}
From this, we observe that the optimization will increase the log-likelihood of the preferred generation (according to the reward model) and decrease that of the dispreferred one, proportionally to the reward difference.

To further simplify the practical implementation, one can easily verify that
\begin{equation}
    \nabla \ell_\copg(y,y';\pi) \propto - \nabla\left(R_\beta^\pi(y)-R_\beta^\pi(y')\right)^2 ,
\end{equation}
We provide the related pseudocode (used in our experiments) in Alg.~\ref{alg:copg}.

\begin{algorithm}
\caption{Practical offline \copg}\label{alg:copg}
\SetKwInOut{Input}{input}
\Input{Dataset $\mathcal{D} = \{(x,y,y')\}$, reference model $\pi_\text{ref}$, model to train $\pi_\theta$}
Initialize $\pi_\theta \leftarrow \pi_\text{ref}$\;
\For{$t\leftarrow 1$ \KwTo $T$}{
    Sample a batch $\mathcal{B}$ from $\mathcal{D}$\;
    Define the regularized reward\;
    $
        R_\beta^\pi(x,y) = R(x,y) - \beta\ln\frac{\pi_\theta(y|x)}{\pi_\text{ref}(y|x)}
    $\;
    Compute the gradient\;
    $
        g_t = \frac{1}{|\mathcal{B}|}\sum_{x,y,y'\in\mathcal{B}}\nabla\left(R_\beta^\pi(x,y) - R_\beta^\pi(x,y')\right)^2
    $\;
    Update the parameters\;
    $
        \theta_{t+1} \leftarrow \theta_t - \eta_t \nabla g_t
    $\;
}
\end{algorithm}

\section{Related works}

Contrastive Policy Gradient is related to policy gradient \citep{williams1991function}. It can be seen as a sound off-policy policy gradient, but crucially not relying on importance sampling, and thus not requiring clipping techniques such as Proximal Policy Optimization \citep{schulman2017proximal} or variations \citep{wu2023pairwise}, allowing for broader applicability (notably, PPO cannot be applied offline to a dataset of unknown density). This link is even stronger:
\begin{prop}[\copg{} and policy gradient]
\label{prop:pg}
    \copg{} generalizes policy gradient in the sense that
    \begin{equation}
        \E_{y\sim\pi, y'\sim\pi}[\nabla  \ell_\copg (y,y';\pi)] = 2 \nabla J(\pi).
    \end{equation}
\end{prop}
The expectation of the gradient of $\ell_\copg$ according to the current policy is exactly (up to the scaling) the policy gradient of Eq.~\eqref{eq:gradient_pg}. So, we can retrieve policy gradient as a special case of the proposed approach, which is much less restrictive as it does not require the generations to be sampled according to the current policy. This result is asymptotic, as the expectation requires an infinite number of generations, but we have a similar connection to a more practical policy gradient approach.
\looseness=-1

Reinforce Leave-One Out is a sample-based policy gradient, using a Monte Carlo estimate of the expected reward from $k$ generations as a baseline \citep{kool2019buy}. It is remarkably effective for finetuning LLMs, simpler than PPO while providing better results, but still relying on fresh generations for each mini-batch \citep{ahmadian2024back}. When using only two generations, the gradient is naturally symmetrized not to waste information and matches exactly Eq.~\eqref{eq:grad_sample_copg}.
\looseness=-1

\begin{prop}[\copg{} and RLOO]
\label{prop:rloo}
    The sample-based gradient $\nabla \ell_\copg (y,y';\pi)$ is exactly the gradient of RLOO for $k=2$, when both $y$ and $y'$ are sampled from the current policy $\pi$.
\end{prop}
A core difference is that \copg{} is valid for any sampling distribution, while RLOO critically relies on using on-policy generations when derived from objective~\eqref{eq:pg_objevctive}. This result shows that it is valid and principled to use the RLOO gradient in an off-policy manner. This is highly non-trivial, new to the community, and made possible thanks to the proposed principled approach. %

Contrastive Policy Gradient can also be related to direct alignment methods \citep{rafailov2023direct,zhao2023slic,azar2024general,tang2024generalized}, and more especially to Identity Policy Optimization \citep{azar2024general}, in the following sense.
\begin{prop}[\copg{} and IPO]
\label{prop:ipo}
    For a pair of generations $(y,y')$, assume without loss of generality that $y$ is preferred to $y'$ according to the reward model, and redefine $R(y) = - R(y') = \frac{1}{4}$, then we have 
    \begin{align}
        &\nabla \ell_\copg (y,y';\pi) =
        \\
        &-\frac{1}{2\beta} \left(\frac{1}{2} - \beta \left(\ln\frac{\pi(y)}{\piref(y)} - \ln\frac{\pi(y')}{\piref(y')}\right)\right)^2,
    \end{align}
    where the term on the right-hand side is the gradient of the sample-based IPO loss to be minimized.
\end{prop}
These results show that if we replace the reward in our objective with a binary signal depending on which generation is preferred, we follow the same gradient as IPO.
In that sense, our approach also subsumes direct alignment approaches, allowing us to optimize for an arbitrary reward.

\section{Toy experiment}
\label{sec:bandit}

\begin{figure}
    \centering
    \includegraphics[width=1\linewidth]{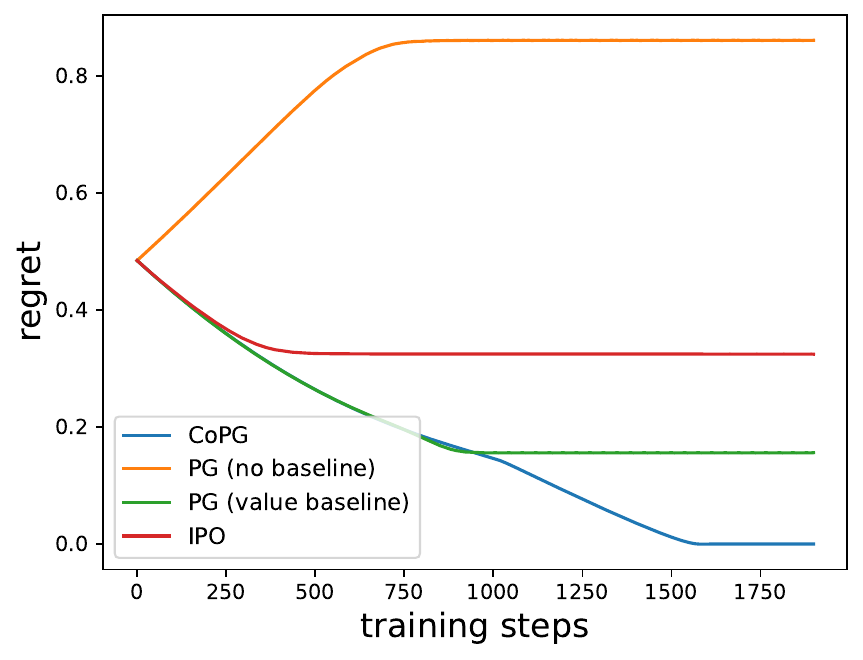}
    \caption{Bandit experiment. \copg{} achieves zero regret, converging to the optimal solution. IPO converges to a biased solution, as it optimizes for the expected preference. %
    PG without a baseline has increasing regret, and PG with a value baseline converges to a biased solution.}
    \label{fig:regret}
\end{figure}

For an illustrative purpose, we consider a simple bandit problem, with 3 arms rewarded by $R = (2.5, 2, 1)$. We choose the data distributions to be $\mu_1 = (0.1, 0.2, 0.7)$ and $\mu_2 = (0.05, 0.05, 0.9)$. Using these distributions, we sample a dataset of $10^4$ pairs of rewarded arms. We set $\beta=0.5$ and  $\piref(y) =\frac{1}{3}$ for $y\in\{1,2,3\}$. The analytical solution to this bandit problem is $\pi_*(y)\propto \exp\frac{R(y)}{\beta}$.

We consider the practical $\copg$ objective of Eq.~\eqref{eq:practical_copg_objective}, for which we recall the gradient Eq.~\eqref{eq:grad_sample_copg}. Given that \copg{} generalizes policy gradient (PG, Prop.~\ref{prop:pg}), we also experiment with it to illustrate the importance of choosing the right baseline. For PG, we consider the following sample-based gradient:
\begin{align}
    \nabla \ell_\text{PG}&(y,y';\pi) = (R^\pi_\beta(y) - b)\nabla\ln\pi(y) 
    \\
    &+ (R^\pi_\beta(y') - b)\nabla\ln\pi(y'). 
\end{align}
As explained before, this gradient is valid whenever both $y$ and $y'$ are sampled according to the current policy $\pi$. However, here we use pairs of arms sampled from the dataset. In other words, this can be seen as a naive off-policy policy gradient. We consider two kind of baselines, $b=0$ (no baseline) and $b=\E_{y\sim\pi}[R(y)]$ (value baseline). The first case corresponds to vanilla policy gradient, and the second case corresponds to the baseline most often used in the literature. Notice that in practice this should be estimated (typically with a value network), but we compute it exactly in this experiment. Given the link between \copg{} and IPO (Prop.~\ref{prop:ipo}), we also experiment with IPO.

For each approach, we train the policy $\hat{\pi}$ with stochastic gradient descent. We use Adam \citep{kingma2014adam} with learning rate $10^{-3}$, batches of size 512, and train for 100 epochs. We measure the performance of the trained policy with the regret:
\begin{equation}
    \text{regret} = J(\pi_*) - J(\hat{\pi}),
\end{equation}
with $J$ being defined Eq.~\eqref{eq:pg_objevctive}.
If \copg{} and PG both rely on a reward function, IPO can only use preferences. We simply set them to be sampled according to a Bradley-Terry Model, $P(y>y') = \sigma(R(y)-R(y'))$, with $\sigma$ the logistic function.

Results are presented in Fig.~\ref{fig:regret}. We can observe that \copg{} converges to the right solution, as predicted theoretically (Thm.~\ref{appx:proofs}). IPO converges to a biased solution. This was to be expected, as it can be shown to optimize the reward $\E_{y'\sim\mu_2}[\sigma(R(y) - R(y')]$ \citep{azar2024general}, which is different from the reward of interest $R$. Regarding policy gradient, we can observe that without baseline, naively applying the policy gradient on off-policy data leads to an increase in the regret: Learning deteriorates the initial policy. Adding the value baseline helps, but it still converges to a biased solution (and it is an ideal algorithm, as here the baseline is analytically computed, while it has to be estimated in practice). Sample-based \copg{} converges to the right solution, showing the importance of choosing the right baseline in an off-policy context. 
\looseness=-1

\section{LLM experiments}
\label{sec:exp_llm}

In this section, we demonstrate the ability of \copg{} to optimize a reward function for finetuning an LLM. As depicted in Sec.~\ref{sec:sample_based_objective}, we consider a pure offline objective, where one has to learn from a fixed dataset of pairs of generations. 
Classic RLHF approaches, such as policy gradient or PPO, do not work in such a pure offline setting. PPO could possibly do a single-step policy improvement, but it would require access to the underlying probabilities $\mu(y)$ of elements in the dataset, which are not available. Moreover, it would require an additional costly value network.
Therefore, as baselines, we consider direct alignment methods, specifically DPO and IPO, for which the preferred completion is chosen according to the same reward model being optimized by \copg.
\looseness=-1

\textbf{Dataset.} We consider the Reddit TL;DR dataset\footnote{\url{https://github.com/openai/summarize-from-feedback}} of \citet{stiennon2020learning}. It is a summarization dataset with an SFT split, consisting of human-written summaries, and a preference split, made of human-annotated preference pairs. We will rerank the preferences according to the reward model, so that \copg{} and direct alignment methods optimize for consistent objectives. 
\looseness=-1

\textbf{Policy Model.} We use Llama2-7B as the base model\footnote{\url{https://huggingface.co/meta-llama/Llama-2-7b-hf}} \citep{touvron2023llama2}. We supervise finetune it on the SFT split of the TL;DR dataset, giving $\piref$ as a result. This is both the initial policy and the reference model for \copg{} as for direct alignment baselines. 
This model is trained for 2 epochs with Adam, with a cosine decay scheduler ($2.10^{-5}$ to 0), warmup of $10\%$, using a batch of size 128.

\textbf{Reward Model.} Our objective is to provide an approach to optimize arbitrary reward functions. As a proof of concept, for this empirical study, we train a reward model using the preference split of the TL;DR dataset and will consider it as the ground truth reward function to be optimized.  We insist right away that we do not claim such a reward model to be the best thing to optimize for improving the LLM, we use it as a proxy for assessing if the proposed approach can indeed optimize a reward at scale. For training the reward function, we use a classic Bradley-Terry model \citep{bradley1952rank}, optimizing for the loss
\begin{equation}
    \ell_\text{RM}(y^+,y^-,R) = -\ln \sigma(R(y^+) - R(y^-)).
\end{equation}
The reward model is trained for two epochs on the train split of the preference dataset, with Adam, the learning rate of $10^{-6}$, a batch of size 128, and a warm-up of $10\%$ of the total number of training steps.
The trained reward model achieves an accuracy of $89.1\%$ on the train set and of $72.8\%$ on the validation set.

\textbf{Training details. } We train all algorithms for two epochs over the train split of the preference dataset. We use a batch of size 128. We optimize the respective losses with Adam, with a learning rate of $10^{-6}$ in all cases. For all approaches we use a warm-up of $10\%$ of all training steps. For \copg{} and DPO we sweep over $\beta\in\{0.01, 0.03, 0.06, 0.1, 0.3, 1\}$. For IPO we sweep over slightly lower values, specifically $\beta\in\{0.003, 0.01, 0.03, 0.06, 0.1, 0.3\}$. 

\textbf{Evaluation.}
Recall that the objective is to know if \copg{} can optimize a reward function by learning offline from a fixed dataset. To evaluate this, every 50 training steps, we perform generations using the trained model on a fixed batch of 128 prompts from the validation dataset and score them using the reward model. We do the same for IPO and DPO, for which we recall that they are trained for preferences according to the reward model, and not according to the original dataset, for a fair comparison, as the reward model is used for evaluation. We also note that the reward model was not trained on the validation set, only on the train set.

\subsection{Can \copg{} optimize a reward?}

\begin{figure}
    \centering
    \includegraphics[width=1\linewidth]{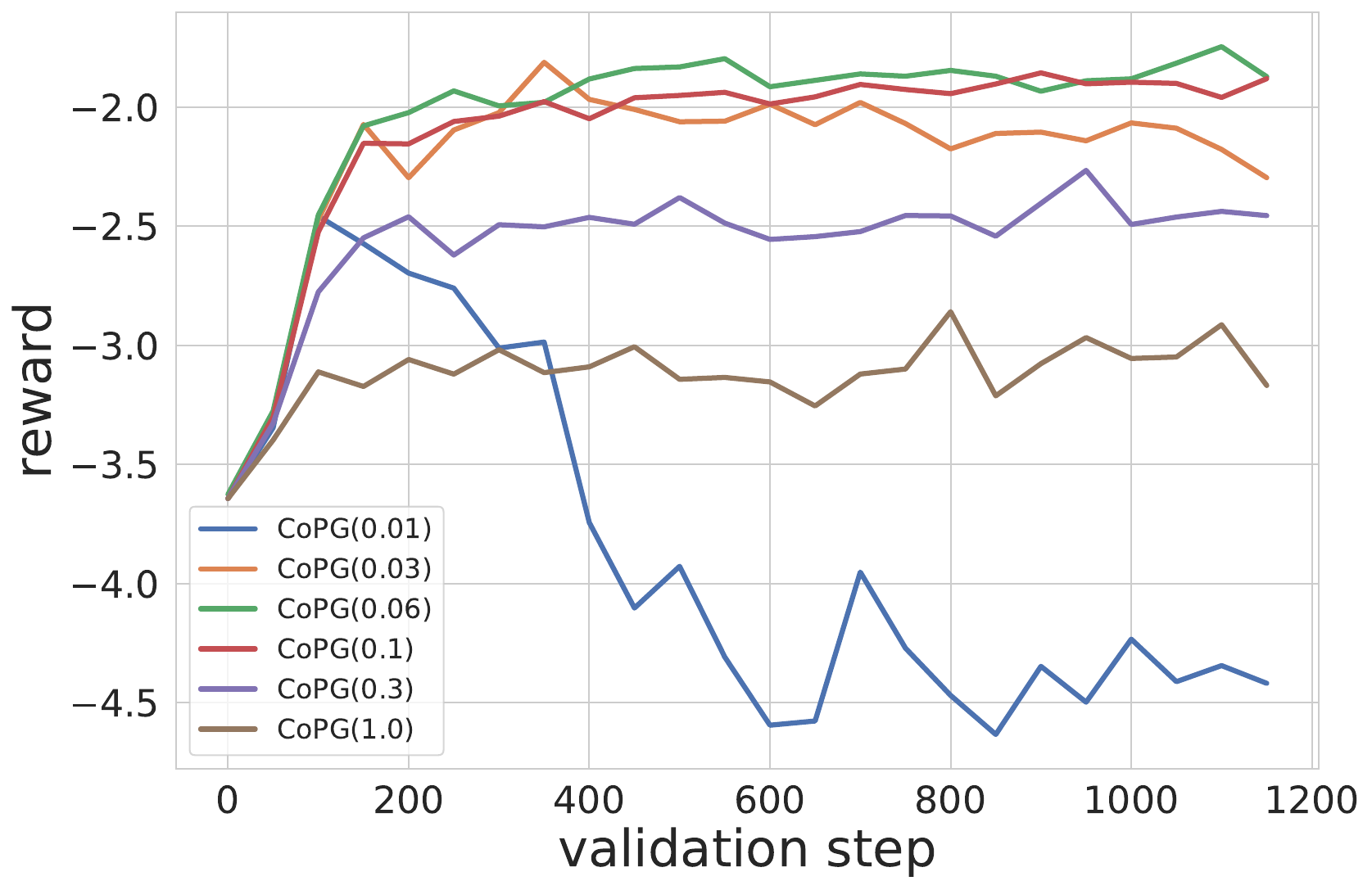}
    \caption{\copg: Rewards of generations along training.}
    \label{fig:copg_rewards}
\end{figure}

\copg{} comes with strong theoretical guarantees, and we here aim to assess its scalability while optimizing a reward in an LLM setting. We train and evaluate it in the setting depicted before, and we provide the corresponding results in Fig.~\ref{fig:copg_rewards}. This figure shows how the reward of generations (averaged over 128 prompts from the validation set) from the model evolves over training (validation each 50 training steps), for the different considered values of $\beta$. 

We can observe that \copg{} successfully optimizes the reward over a large range of temperature $\beta \in [0.03,0.1]$, with the higher reward being achieved for $\beta=0.06$.  
When the temperature is too low, it becomes unstable, and the reward drops. Interestingly, this does not translate into a sign of overfitting in other validation metrics, such as the loss. This is not something new to the LLM community, but it highlights the necessity of doing generations when evaluating a model, which is a classic RL thing, especially in a pure offline setting. 
When the temperature is too high, the reward still increases, but to a lower value. This has to be expected. In this case, the Kullback-Leibler term becomes predominant, and the policy is incentivized more to avoid moving too far away from the reference model, which was also the initial policy.
\looseness=-1

\begin{figure}
    \centering
    \includegraphics[width=1\linewidth]{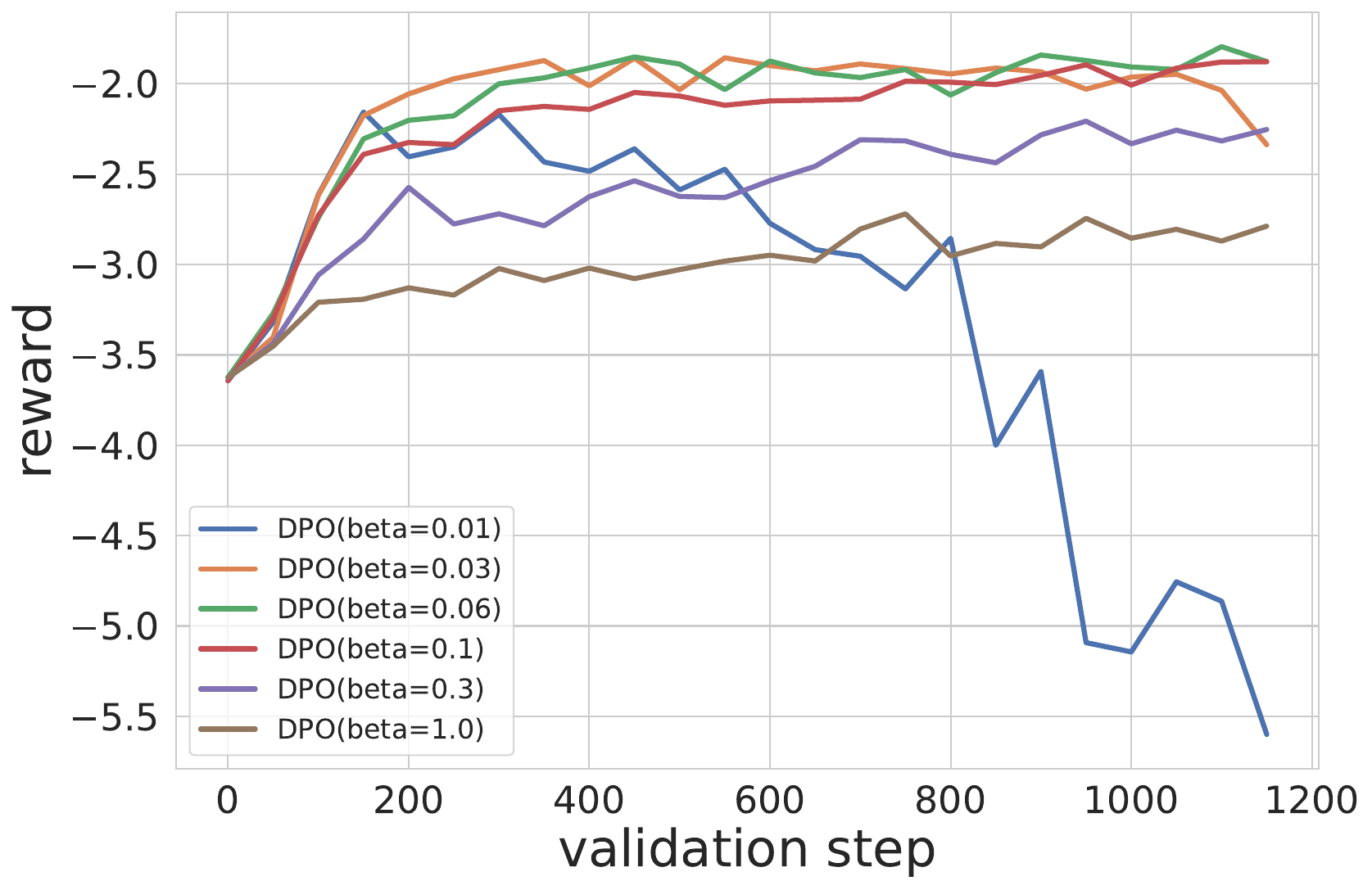}
    \caption{DPO: Rewards of generations along training.}
    \label{fig:dpo_rewards}
\end{figure}

For a more complete study, we also provide the related results for IPO and DPO. They do not directly aim at optimizing the reward but rather at optimizing the preferences. However, given that, in our case, these preferences are ranked according to the reward model, they should also generate sequences of increasing rewards. For DPO, given that in this specific case the preference follows a Bradley-Terry model, it should indeed optimize the reward function, while IPO optimizes for a different objective, see \citet[Prop.~1 and Thm.~1]{azar2024general} for more details.
\looseness=-1

We provide the result for DPO in Fig.~\ref{fig:dpo_rewards}. We can observe that DPO is indeed able to optimize the reward too. As \copg, it is not too sensitive to the value of $\beta$, for the same range. Similarly, it becomes unstable when $\beta$ is too low, and it increases less the reward when the temperature is too high because it stays closer to the reference model.

\begin{figure}
    \centering
    \includegraphics[width=1\linewidth]{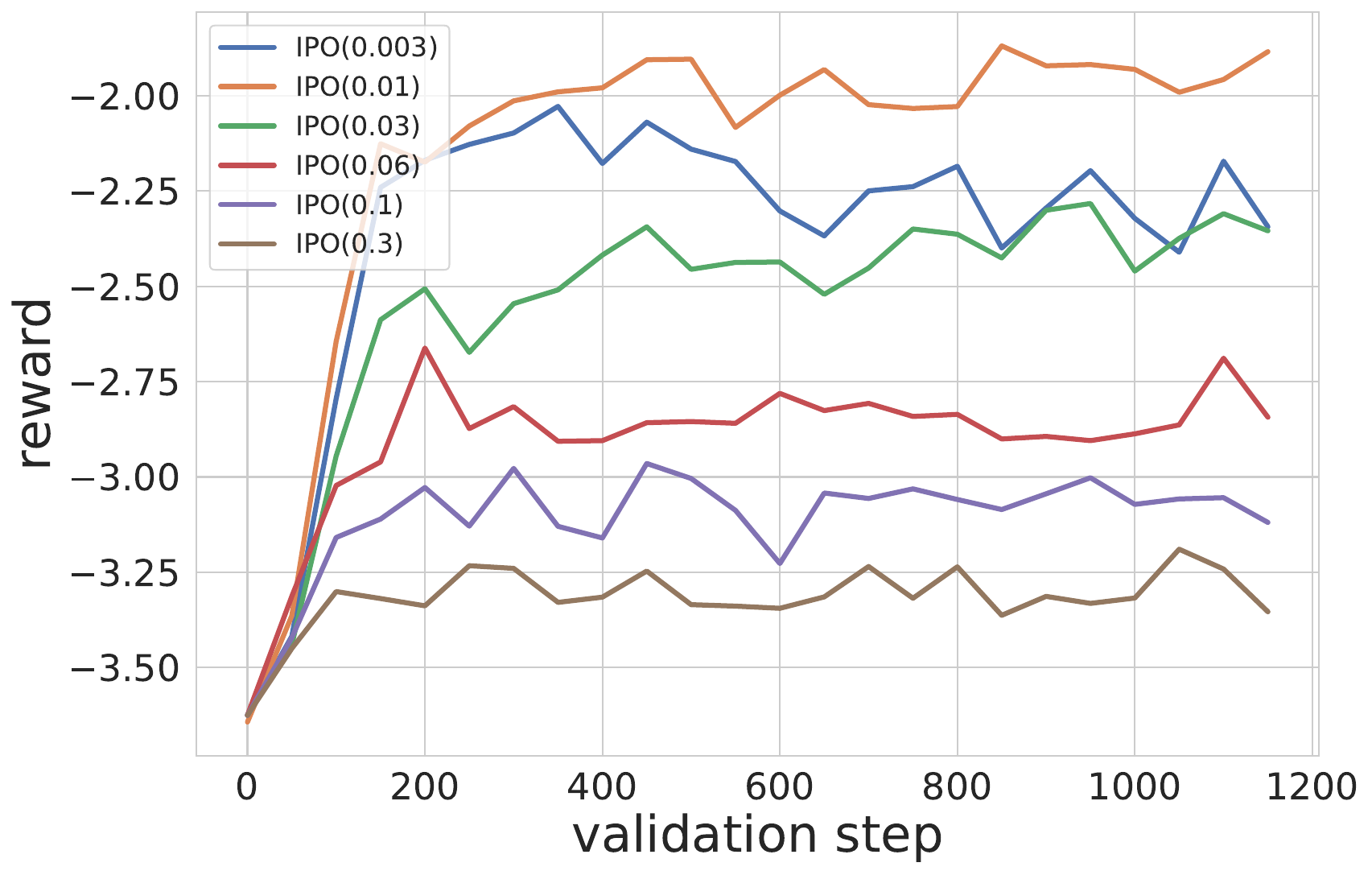}
    \caption{IPO: Rewards of generations along training.}
    \label{fig:ipo_rewards}
\end{figure}

The results for IPO are provided in Fig.~\ref{fig:ipo_rewards}. IPO, too, increases the reward. Conversely to \copg{} or DPO, it seems to be more stable, because we do not observe a significant drop when $\beta$ becomes smaller (but we expect this to happen for lower values of $\beta$). It also appears to be not too sensitive to the value of $\beta$ in a given range (taking into account the difference of scale without a ``dropping'' run), yet for lower values.
 
\begin{figure}
    \centering
    \includegraphics[width=1\linewidth]{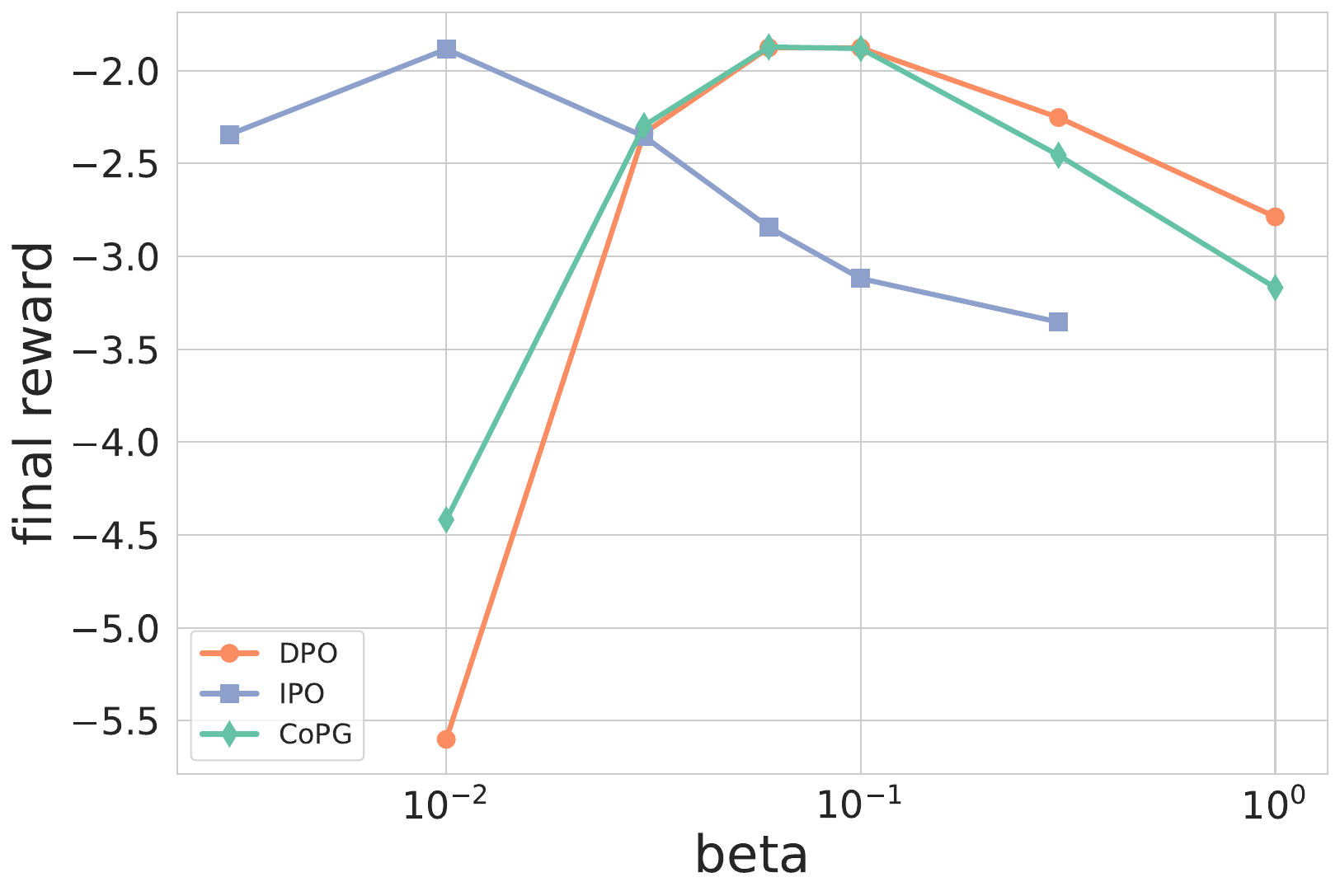}
    \caption{Final reward as a function of $\beta$.}
    \label{fig:reward-beta}
\end{figure}

To summarize these results, we show in Fig.~\ref{fig:reward-beta} the expected reward after training as a function of $\beta$. This showcases the stable range and the fact that the various approaches rely on different ranges of temperature values to provide high rewards.
\looseness=-1

\subsection{How does \copg{} compare to direct alignment?}

So far, we have shown that both \copg{} and the direct alignment methods DPO and IPO were able to increase the reward in an offline manner. However, a core question is to know if directly optimizing a reward, as \copg{} does, provides better results than optimizing for a preference based on this reward function. In a simple and controlled case, such as the bandit experiment of Sec.~\ref{sec:bandit}, the answer is clear, because we can exhibit the optimal solutions and we know to what each method should converge. However, it is much less clear in a large-scale problem, such as an LLM generating sequences.

\begin{figure}
    \centering
    \includegraphics[width=1\linewidth]{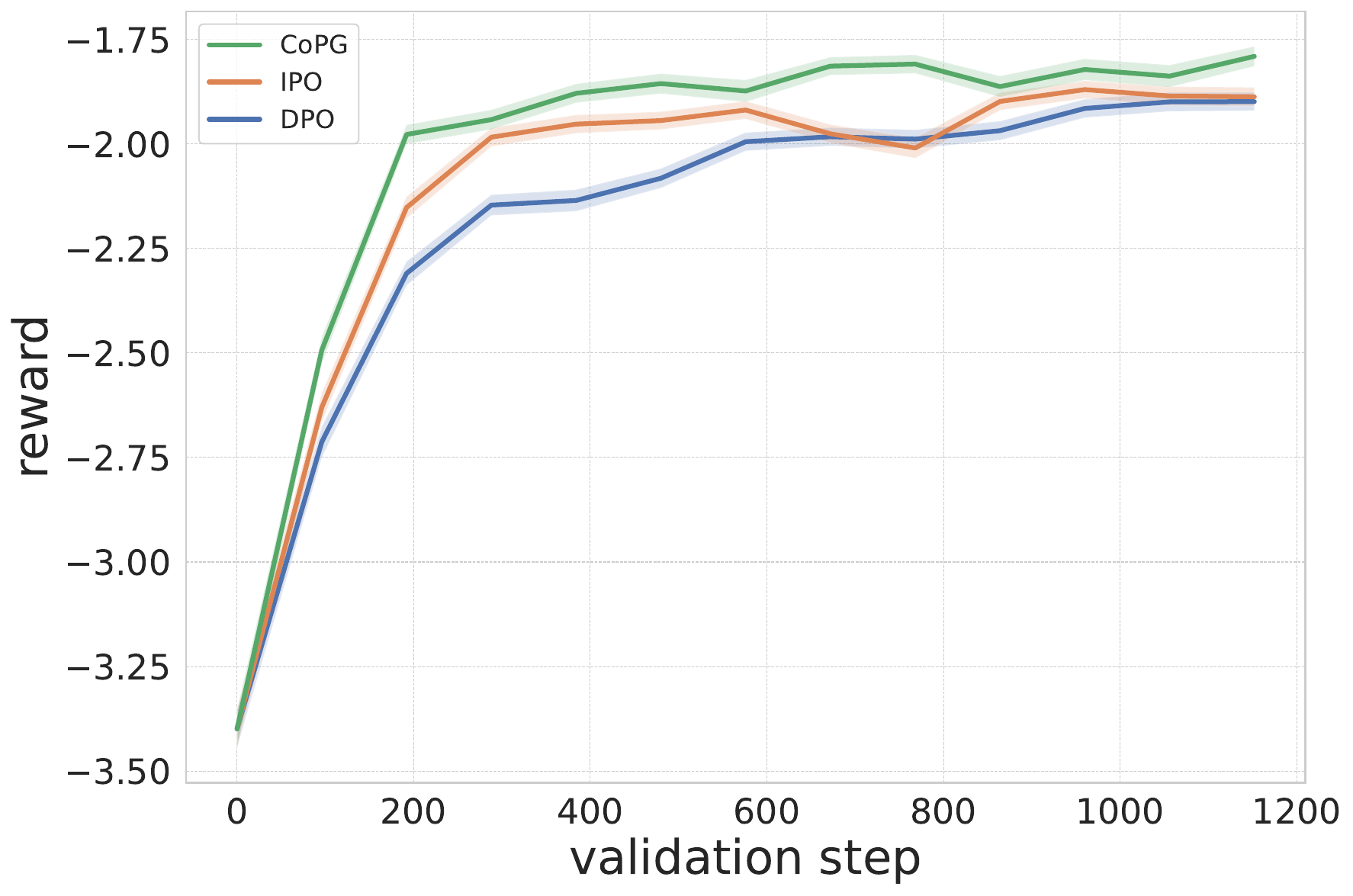}
    \caption{Comparison of \copg, DPO and IPO.}
    \label{fig:compare_best}
\end{figure}

To assess this, we compare \copg{}, DPO and IPO with the best temperature from Fig.~\ref{fig:reward-beta} (respectively $\beta=0.06$, $\beta=0.1$ which is also the classic value for DPO in the literature, and $\beta=0.01$). We also rerun the experiments, this time gathering generations for a batch of 1024 prompts from the validation set, to get a better estimate of the expected reward, doing this each 100 training steps.

Results are presented in Fig.~\ref{fig:compare_best}, the shaded envelop corresponding to the standard errors. We can observe that \copg{} consistently achieves a higher reward, and faster, than both IPO and DPO. We hypothesize that it could be even more true for a reward not trained from preference data.

In this case, the preferences following the Bradely-Terry model (because ranked according to the reward function), DPO should optimize for the reward function, according to \citet[Prop.~1]{azar2024general}. Yet, it achieves the lowest reward in this experiment. We think that this is due to this convergence result being an asymptotical one. In practice, \copg{} uses explicitly the reward signal, while DPO only uses a binarized signal (which completions is preferred), and it would require the algorithm to observe multiple rankings of the same completions to converge to the right solution. 

In principle, IPO converges to a different solution, but for the same $\beta$. For the stated theoretical results, $\beta$ is part of the problem, while in practice it is a hyperparameter. Here, we have chosen $\beta$ so as to achieve the highest possible reward, and the effective $\beta$ for IPO is much smaller than for both \copg{} and IPO. However, IPO also achieves a lower reward than \copg, even if closer than DPO.
\looseness=-1

Overall, this experiment also suggests that if we are given a reward function to optimize, it may not be sufficient to use it to build a preference dataset with preferences ranked according to the reward model so as to learn from it using a direct alignment method. Reinforcement learning approaches still have a place in this field, and our proposed \copg{} allows to maximize efficiently the reward, in a purely off-policy manner, while being as simple, stable and computationally lightweight as direct alignment approaches.

\subsection{KL-Reward trade-off}

In regularized RL, the objective is to maximize the reward while keeping the KL divergence between the policy and the reference model small enough. This is especially important for LLMs, as this can prevent reward hacking for example. For this experiment, we consider the HH dataset from \citet{bai2022training}. We use the same recipe as TL;DR for training the SFT checkpoint (this time using the preferred examples, in the absence of SFT data), the reward model, IPO, DPO and \copg. We also use the same hyperparameters. 

\begin{figure}
    \centering
    \includegraphics[width=1\linewidth]{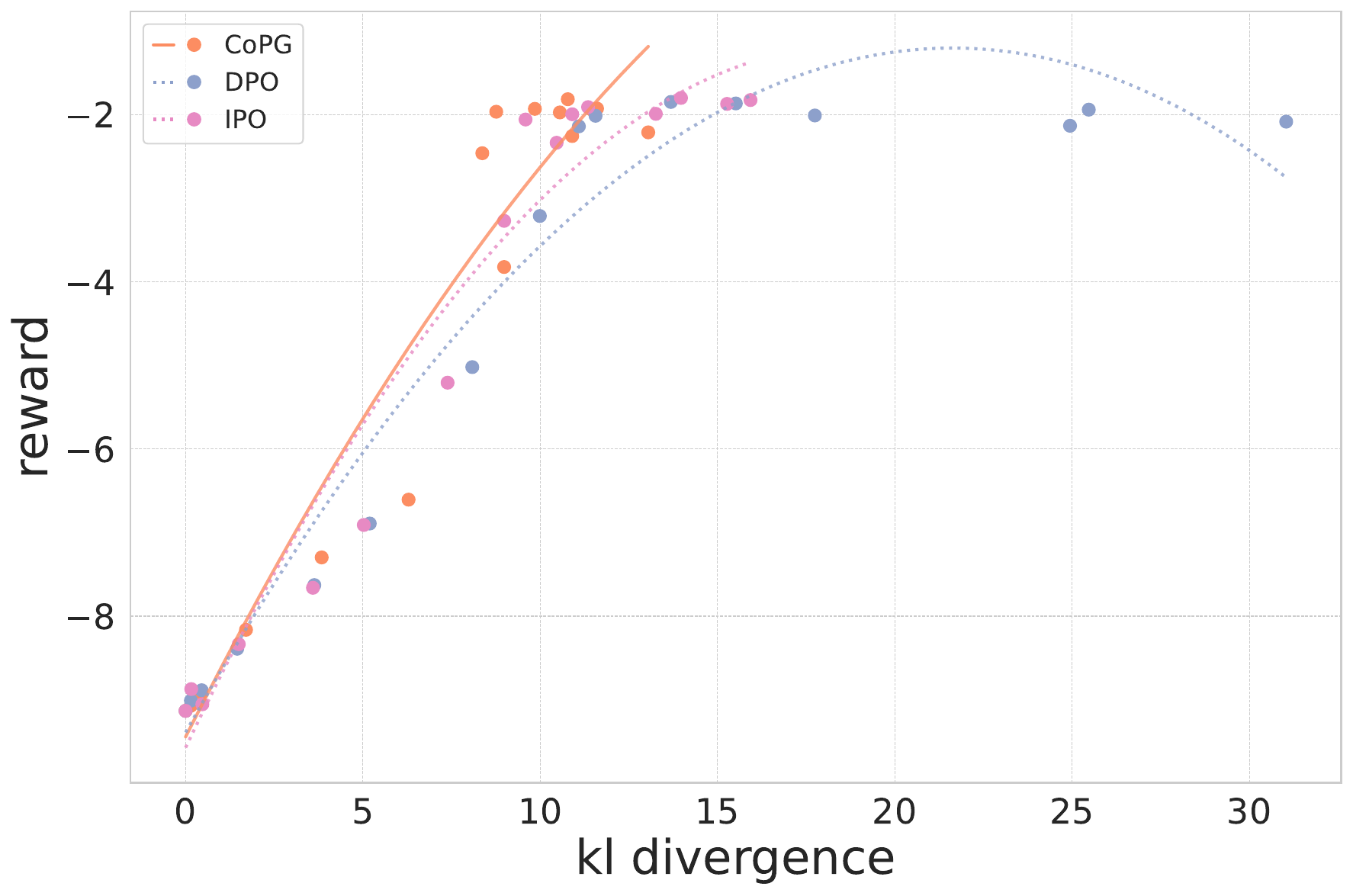}
    \caption{KL-Reward trade-off for \copg, DPO and IPO.}
    \label{fig:kl-reward}
\end{figure}

We follow the same methodology, sweeping for each algorithm over values of $\beta$, selecting the one with the higher final reward averaged of 128 generations. Then, we retrain each algorithm with its best value of $\beta$, using 1024 generations for computing the reward at each validation step. From that, we compute the KL/reward Pareto front for IPO, DPO and \copg.

Results are presented in Fig.~\ref{fig:kl-reward}. It is a scatter plot of the rewards as a function of the KL divergences, each point corresponding to one of the 1024 generations for a specific validation step and a specific algorithm.  The figure also provides a polynomial interpolation. The Pareto front clearly shows that \copg outperforms IPO and DPO in achieving higher rewards while maintaining lower KL divergence. In particular, we note that DPO tends to achieve a higher KL more easily than other methods.

\section{Discussion and perspectives}

We have introduced \emph{Contrastive Policy Gradient}, a new RL approach for finetuning LLMs. It is a form of policy gradient that contrasts the reward with a specific baseline. The corresponding objective function is supervised-friendly, in the sense that it does not (necessarily) rely on fresh generations from the model. This allows to learn a policy in a pure offline setting, without relying on importance sampling or clipping of log-probability ratios, and does not require the introduction of an additional value network. We have demonstrated that \copg{} indeed optimizes for the optimal KL-regularized policy (Thm.~\ref{thm:right_solution}), and we have shown that it generalizes policy gradient (Prop.~\ref{prop:pg}), RLOO (Prop.~\ref{prop:rloo}), or IPO (Prop.~\ref{prop:ipo}).

On a controlled but simple bandit experiment (Sec.~\ref{sec:bandit}), we have illustrated empirically the convergence properties of the proposed approach, the importance of choosing the right baseline (the one coming from our derivations rather than the classic value baseline, and even less no baseline), and the advantage of optimizing a reward function rather than preferences derived from these rewards, which leads to a biased solution. On a larger scale LLM experiment (Sec.~\ref{sec:exp_llm}), we have shown that \copg{} is able to optimize a reward function, in a fully offline and off-policy manner, conversely to other RL-finetuning approaches, and that it can achieve a higher reward than by using a direct alignment approach on preferences ranked by the reward model. 
Our experimental study also suggests that this higher reward is achieved at a lower KL cost, compared to IPO or DPO.

A core perspective is to study \copg{} in an online setting. Indeed, if it works in the offline setting, being the first such RL approach in the context of LLMs (to the best of our knowledge), it is not restricted to this setting. Recent works have highlighted the benefits of using online (or fresh) data for direct alignment \citep{tang2024understanding,tajwar2024preference}, and we hypothesize that these findings can also benefit to \copg. Typically, one can consider using a replay buffer, as classically done in off-policy RL but less common for LLM finetuning. As \copg{} uses pairs of generations, we think that this can also open new perspectives by using heterogeneous distributions for the completions, as briefly discussed in Sec.~\ref{sec:sample_based_objective}. For example, one distribution could correspond to the replay buffer, while the other could correspond to exploratory generations (for addressing the exploration/exploitation dilemma), or by using a dataset of good but suboptimal generations, in the spirit of RL from demonstrations. Another important perspective is to experiment \copg{} on more tasks and rewards, and we also plan to study its possible extension to the multi-objective RL setting.

\section*{Limitations}

If the proposed \copg{} approach comes with strong theoretical guarantees and has been validated in both a simple bandit problem and a larger scale LLM experiments, it would benefit from being assessed on more tasks and rewards in the context of LLMs. \copg{} works in a pure offline setting, which is a strength, but it would benefit from using fresh generations too, as well as from possibly heterogeneous sources of data. Nothing prevents it in principle, but this has still to be investigated, and would provide a fair comparison to other classic RL finetuning approaches, relying on (near) on-policy samples. The proposed approach optimizes for a single reward model, its extension to multiple rewards remains an interesting open question. Also, our approach assumes that the reward model is reliable, which is often not the case in practice, especially when it is learnt from data. This is in part the role of KL-regularization (to avoid hacking the reward), but our approach has no additional mechanism for preventing optimizing bad areas of the reward model. This would be especially important for a number of LLM applications.

\bibliography{custom}

\newpage

\onecolumn

\appendix

\section{Proofs of theoretical results}
\label{appx:proofs}
\setcounter{theorem}{0}
\setcounter{prop}{0}

In this section, we prove the stated theoretical results. First, we reintroduce the notations, now taking into account the prompt $x$ (or context). The regularized reward defined in Eq.~\eqref{eq:reg_reward} becomes 
\begin{equation}
    R_\beta^\pi(x,y) = R(x,y) - \beta\ln\frac{\pi(y|x)}{\piref(y|x)}.
    \label{eq:reg_reward_full}
\end{equation}
The classic RL problem of Eq.~\eqref{eq:pg_objevctive} writes
\begin{equation}
    J(\pi) = \E_{x\sim\rho, y\sim\pi(\cdot|x)}[R(x,y) - \beta \kl{\pi(\cdot|x)}{\piref(\cdot|x)}] = \E_{x\sim\rho,y\sim\pi(\cdot|x)}[R_\beta^\pi(x,y)].
    \label{eq:full_rl_objective}
\end{equation}
The unique maximizer of $J$ is well known, $J$ being a Legendre-Fenchel transform (\textit{e.g.}, see \citet[Appx.~A]{vieillard2020leverage}), it is given by
\begin{equation}
    \pi_*(y|x) = \frac{\piref(y|x)\exp\frac{R(x,y)}{\beta}}{Z_*(x)},
    \label{eq:optimal_policy}
\end{equation}
with $Z_*(x) = \sum_y \piref(y|x) \exp\frac{R(x,y)}{\beta}$ the associated partition function.

The proposed \copg{} loss \eqref{eq:loss_copg} with context $x$ simply writes
\begin{equation}
    \ell_\copg(x,y,y';\pi) = \left(R_{\beta/2}^\pi(x,y) - R_{\beta/2}^\pi(x,y')\right) \ln \frac{\pi(y|x)}{\piref(y|x)}
    + \left(R_{\beta/2}^\pi(x,y') - R_{\beta/2}^\pi(x,y)\right) \ln \frac{\pi(y'|x)}{\piref(y'|x)}.
\end{equation}
The associated objective function \eqref{eq:copg_objective_function} to be maximized is then, with context $x$,
\begin{equation}
    L(\pi) = \E_{x\sim\rho, y\sim\mu_1(\cdot|x),y'\sim\mu_2(\cdot|x)}[\ell_\copg(x,y,y';\pi)].
    \label{eqappx:copg_objective}
\end{equation}

Now, we restate Thm.~\ref{thm:right_solution} in a more general form and prove it.
\begin{theorem}[\copg{} solves the right problem.]
    Assume that $\piref$, $\mu_1$ and $\mu_2$ all have the same support (that is, for any triplet $(x,y,y')$ such that $\rho(x)>0$, we have $\piref(y|x)>0 \Leftrightarrow \mu_1(y|x)>0 \Leftrightarrow \mu_2(y|x)>0$).
    Then, the unique maximizer of $L(\pi)$, Eq.~\eqref{eqappx:copg_objective}, is the optimal policy $\pi_*$ of Eq.~\eqref{eq:optimal_policy}, which is also the unique maximizer of $J(\pi)$, Eq.~\eqref{eq:full_rl_objective}.
\end{theorem}
\begin{proof}
    We start by showing that $\pi_*$ is a maximizer, before proving that it is the sole one. First recall the \copg{} loss:
    \begin{equation}
        \ell_\copg(x,y,y';\pi) = \left(R_{\beta/2}^\pi(x,y) - R_{\beta/2}^\pi(x,y')\right) \ln \frac{\pi(y|x)}{\piref(y|x)}
        + \left(R_{\beta/2}^\pi(x,y') - R_{\beta/2}^\pi(x,y)\right) \ln \frac{\pi(y'|x)}{\piref(y'|x)}.
    \end{equation}
    Without loss of generality, thanks to the support assumption, we can reparametrize the policy $\pi$ as follows:
    \begin{equation}
        \beta \ln \frac{\pi(y|x)}{\piref(y|x)} = V(x,y) - Z_V(x),
        \label{eq:reparametrization}
    \end{equation}
    with $Z_V(x) = \beta \ln \sum_y \piref(y|x)\exp\frac{V(x,y)}{\beta}$ the associated (scaled) partition function. In essence, $V(x,y)$ can be understood as the logits of the learnt policy, shifted by the log-probabilites of the reference policy.
    
    Then, we can rewrite the \copg{} loss using the above reparametrization:
    \begin{align}
        &\beta \ell_\copg(x,y,y';V)
        \\
        &\stackrel{\textit{(a)}}{=} 
        \left(R(x,y) - \frac{1}{2} (V(x,y) - Z_V(x)) - R(x,y') + \frac{1}{2} (V(x,y') -  Z_V(x)) \right) \left(V(x,y) -  Z_V(x)\right)
        \\
        &\; + \left(R(x,y') - \frac{1}{2} (V(x,y') - Z_V(x)) - R(x,y) + \frac{1}{2} (V(x,y) -  Z_V(x)) \right) \left(V(x,y') -  Z_V(x)\right)
        \\
        &\stackrel{\textit{(b)}}{=} 
        \left(R(x,y) - R(x,y') - \frac{1}{2}(V(x,y) - V(x,y')) \right)\left(V(x,y) - V(x,y')\right)
        \\
        &\stackrel{\textit{(c)}}{=}
        \frac{1}{2} (R(x,y) - R(x,y'))^2 - \frac{1}{2}\left(R(x,y) - R(x,y') - (V(x,y) - V(x,y'))\right)^2. \label{eq:proof_square_term}
    \end{align}
    In the above derivations, $\textit{(a)}$ is true by using the reparametrization of Eq.~\eqref{eq:reparametrization}, $\textit{(b)}$ is obtained by canceling terms (all terms $Z_V(x)$ are weighted by $0$) and refactoring, and $\textit{(c)}$ is easily obtained by recoginizing a partial square expansion in $\textit{(b)}$ (of the form $\frac{1}{2}(\Delta V)^2 - (\Delta V)(\Delta R)$).

    Hence, a pointwise maximizer of $\ell_\copg(x,y,y';V)$ is necessarily a minimizer of $(R(x,y) - R(x,y') - (V(x,y) - V(x,y')))^2$ (the term $(R(x,y) - R(x,y'))^2$ being constant with respect to optimization), and $V=R$ is obviously such a minimizer, setting the square term to $0$. With $V=R$, Eq.~\eqref{eq:reparametrization} characterizes the optimal policy $\pi_*$ of Eq.~\eqref{eq:optimal_policy}. Therefore, we have just shown that $\pi_*$ is a maximizer of $L(\pi)$.

    Now, let us show that this maximizer is unique. Let $\tilde{\pi}$ be a maximizer of $L(\pi)$, and let $\tilde{V}$ be an associated logit function according to Eq.~\eqref{eq:reparametrization} (notice that there is no unicity of the logits, a shift by an $x$-dependant function provides the same equation). The term $\tilde{V}$ necessarily sets the square term of Eq.~\eqref{eq:proof_square_term} to zero (because $V=R$ does so). Therefore, for any triplet $(x,y,y')$ such that $\rho(x)>0$, $\piref(y|x)>0$ and $\piref(y'|x)>0$, we have that 
    \begin{equation}
        R(x,y) - \tilde{V}(x,y) = R(x,y') - \tilde{V}(x,y'). 
    \end{equation}
    This is not enough to ensure unicity, $\tilde{V}(x,y) - b(x)$ would satisfy this equality for an arbitrary $b(x)$. However, we're interested in the policy solution.
    We have that:
    \begin{align}
        &\quad R(x,y) - \tilde{V}(x,y) = R(x,y') - \tilde{V}(x,y')
        \\
        \stackrel{\textit{(a)}}{\Leftrightarrow}
        &\quad R(x,y) - \beta\ln\frac{\tilde{\pi}(y|x)}{\piref(y|x)} = R(x,y') - \beta\ln\frac{\tilde{\pi}(y'|x)}{\piref(y'|x)} 
        \\
        \stackrel{\textit{(b)}}{\Leftrightarrow}
        &\quad \beta \ln \pi_*(y|x) - \beta \ln \tilde{\pi}(y|x) = \beta \ln \pi_*(y'|x) - \beta \ln \tilde{\pi}(y'|x) 
        \\
        \stackrel{\textit{(c)}}{\Leftrightarrow}
        &\quad \pi_*(y'|x) = \frac{\pi^*(y|x) \tilde{\pi}(y'|x)}{\tilde{\pi}(y|x)}
        \\
        \stackrel{\textit{(d)}}{\Rightarrow}
        &\quad 1 = \sum_{y'} \pi_*(y'|x) = \sum_{y'} \frac{\pi^*(y|x) \tilde{\pi}(y'|x)}{\tilde{\pi}(y|x)} = \frac{\pi^*(y|x)}{\tilde{\pi}(y|x)}
        \\
        \Leftrightarrow
        &\quad \tilde{\pi}(y|x) = \pi_*(y|x).
    \end{align}
    In the above derivation, \textit{(a)} is true by Eq.~\eqref{eq:reparametrization} and canceling the terms $\ln Z_{\tilde{V}}(x)$ appearing in both sides, \textit{(b)} is true by recognizing from Eq.~\eqref{eq:optimal_policy} that  $\beta\ln\piref(y|x) + R(x,y) = \beta \ln\pi_*(y|x) - \beta \ln Z_*(x)$ and canceling the terms $\beta \ln Z_*(x)$ appearing on both sides, \textit{(c)} is true by simplifying $\beta$, exponentiating and rearranging, and \textit{(d)} is true by using the fact that both $\pi_*(\cdot|x)$ and $\tilde{\pi}(\cdot|x)$ are distributions.

    We have just shown that any maximizer $\tilde{\pi}$ of $L$ is necessarily $\pi_*$, which concludes the proof.
\end{proof}

Next, we restate Property~\ref{prop:pg} and prove it.

\begin{prop}[\copg{ and policy gradient}]
    \copg{} generalizes policy gradient in the sense that 
    \begin{equation}
        \E_{x\sim\rho, y\sim\pi(\cdot|x), y\sim\pi(\cdot|x)}[\nabla \ell_\copg(x,y,y';\pi)] = 2 \nabla J(\pi).
    \end{equation}
\end{prop}
\begin{proof}
    Let start by reproving the classic policy gradient. We have that
    \begin{align}
        \nabla J(\pi) &= \nabla \E_{x\sim\rho,y\sim\pi(\cdot|x)}[R^\pi_\beta(x,y)]
        \\
        &=  \E_{x\sim\rho,y\sim\pi(\cdot|x)}[R^\pi_\beta(x,y) \nabla \ln \pi(y|x) + \nabla R^\pi_\beta(x,y)]
        \\
        &= \E_{x\sim\rho,y\sim\pi(\cdot|x)}[R^\pi_\beta(x,y) \nabla \ln \pi(y|x)],
    \end{align}
    where for the last step we make use of the fact that $\E_{x\sim\rho, y\sim\pi(\cdot|x)}[\nabla \ln\pi(y|x)] = 0$.

    Now, let compute the gradient of the \copg{} loss:
    \begin{align}
        &\nabla \ell_\copg(x,y,y';\pi)
        \\
        &= \nabla \left(
        \left(R_{\beta/2}^\pi(x,y) - R_{\beta/2}^\pi(x,y')\right) \ln \frac{\pi(y|x)}{\piref(y|x)}
        + \left(R_{\beta/2}^\pi(x,y') - R_{\beta/2}^\pi(x,y)\right) \ln \frac{\pi(y'|x)}{\piref(y'|x)}
        \right)
        \\
        &= \nabla \left(R_{\beta/2}^\pi(x,y) - R_{\beta/2}^\pi(x,y')\right) \ln \frac{\pi(y|x)}{\piref(y|x)}
        + \left(R_{\beta/2}^\pi(x,y) - R_{\beta/2}^\pi(x,y')\right) \nabla \ln \frac{\pi(y|x)}{\piref(y|x)}
        \\
        &\; + \nabla \left(R_{\beta/2}^\pi(x,y') - R_{\beta/2}^\pi(x,y)\right) \ln \frac{\pi(y'|x)}{\piref(y'|x)}
        +\left(R_{\beta/2}^\pi(x,y') - R_{\beta/2}^\pi(x,y)\right) \nabla \ln \frac{\pi(y'|x)}{\piref(y'|x)}
        \\
        &= \left(R_{\beta}^\pi(x,y) - R_{\beta}^\pi(x,y')\right) \nabla \ln \pi(y|x)
        + \left(R_{\beta}^\pi(x,y') - R_{\beta}^\pi(x,y)\right) \nabla \ln \pi(y'|x).
        \label{eq:proof:grad_copg}
    \end{align}
    It is important to not ignore the fact that $\R^\pi_{\beta/2}$ does depend on $\pi$, and thus contributes to the gradient, the rest of derivations skipped above are simple calculus and rearranging terms.

    So, the gradient $\nabla \ell_\copg(x,y,y';\pi)$ is a sum of two terms, let focus on the first one. We have that
    \begin{align}
        &\E_{x\sim\rho, y\sim\pi(\cdot|x), y'\sim\pi(\cdot|x)}[(R_{\beta}^\pi(x,y) - R_{\beta}^\pi(x,y')) \nabla \ln \pi(y|x)]
        \\
        = &\E_{x\sim\rho, y\sim\pi(\cdot|x)}[R_{\beta}^\pi(x,y) \nabla \ln \pi(y|x)]
        -\E_{x\sim\rho}\left[
        \E_{y'\sim\pi(\cdot|x)}[R_{\beta}^\pi(x,y')] \underbrace{\E_{y\sim\pi(\cdot|x)}[\nabla\ln\pi(y|x)]}_{=0}
        \right]
        \\
        = &\nabla J(\pi).
    \end{align}
    By symmetry, we have exactly the same result for the second term,
    \begin{equation}
        \E_{x\sim\rho, y\sim\pi(\cdot|x), y'\sim\pi(\cdot|x)}[(R_{\beta}^\pi(x,y') - R_{\beta}^\pi(x,y)) \nabla \ln \pi(y'|x)] = \nabla J(\pi),
    \end{equation}
    which overall shows that 
    \begin{equation}
        \E_{x\sim\rho, y\sim\pi(\cdot|x), y'\sim\pi(\cdot|x)}[\nabla \ell_\copg(x,y,y';\pi)] = 2 \nabla J(\pi).
    \end{equation}
    This concludes the proof.
\end{proof}

Then, we restate Prop.~\ref{prop:rloo} and prove it.

\begin{prop}[\copg{} and RLOO]
    The sampled-based gradient $\nabla \ell_\copg(x,y,y';\pi)$ is exactly the gradient of RLOO for k=2, when both $y$ and $y'$ are sampled from the current policy $\pi$.
\end{prop}
\begin{proof}
    First, recall the gradient of the \copg{} loss from Eq.~\eqref{eq:proof:grad_copg}, proven in the previous proof:
    \begin{equation}
        \nabla \ell_\copg(x,y,y';\pi) = \left(R_{\beta}^\pi(x,y) - R_{\beta}^\pi(x,y')\right) \nabla \ln \pi(y|x)
        + \left(R_{\beta}^\pi(x,y') - R_{\beta}^\pi(x,y)\right) \nabla \ln \pi(y'|x).
    \end{equation}
    Next, we rederive RLOO from first principle.
    Recall the classic policy gradient:
    \begin{equation}
        \nabla J(\pi) = E_{x\sim\rho, y\sim\pi(\cdot|x)}[R_\beta^\pi(x,y) \nabla \ln\pi(y|x)].
    \end{equation}
    A sample-based gradient is given by, with $y$ being sampled according to $\pi(\cdot|x)$,
    \begin{equation}
        \hat{\nabla} J(\pi) = R_\beta^\pi(x,y) \nabla \ln\pi(y|x).
    \end{equation}
    As explained before, a baseline $b(x)$ can be considered, without biasing the gradient, as long as it is independent from the generation $y$:
    \begin{equation}
        \hat{\nabla}_b J(\pi) = (R_\beta^\pi(x,y) - b(x)) \nabla \ln\pi(y|x).
    \end{equation}
    It is easy to check that the gradient is unbiased:
    \begin{equation}
        \E_{y\sim\pi(\cdot|x)} \hat{\nabla}_b J(\pi) = \underbrace{\E_{y\sim\pi(\cdot|x)}[ R_\beta^\pi(x,y) \nabla \ln\pi(y|x)]}_{=\nabla J(\pi)} - b(x) \underbrace{\E_{y\sim\pi(\cdot|x)}[\nabla \ln\pi(y|x)]}_{=0} = \nabla J(\pi).
    \end{equation}
    The principle of RLOO is to perform $k$ independent generations $y^1,\cdots,y^k$ for each prompt $x$, using the current policy $\pi(\cdot|x)$, and to use as a stochastic baseline for $R_\beta^\pi(x,y^j)$, more specifically the leave-one-out empirical expectation of the reward using the $k-1$ other generations. This is still a valid baseline (derivation above applies), as even if the baseline is stochastic, it is independent from $y^j$. The corresponding empirical gradient is
    \begin{equation}
        \hat{\nabla}_k J(\pi) = \sum_{j=1}^k \Big(R_\beta^\pi(x,y^j) - \frac{1}{k-1}\sum_{\substack{l=1\\l\neq j}}^k  R_\beta^\pi(x,y^l)  \Big) \nabla \ln \pi(y^j|x).
    \end{equation}
    In the case $k=2$ this simplifies to:
    \begin{equation}
        \hat{\nabla}_{k=2} J(\pi) =\left(R_{\beta}^\pi(x,y^1) - R_{\beta}^\pi(x,y^2)\right) \nabla \ln \pi(y^1|x)
        + \left(R_{\beta}^\pi(x,y^2) - R_{\beta}^\pi(x,y^1)\right) \nabla \ln \pi(y^2|x).
    \end{equation}
    This is exactly the gradient $\nabla \ell_\copg(x,y,y';\pi)$, which proves the result. However, as explained in the main text, it is crucial to note that RLOO derivation is only valid when generations are done with the current policy, while \copg{} can account for arbitrary generations. In this sense, \copg{} says that RLOO can be safely used in an off-policy context.
\end{proof}

Eventually, we restate and prove Prop.~\ref{prop:ipo}.

\begin{prop}[\copg{} and IPO]
    For a prompt $x$ and a pair of generations $(y,y')$, assume without loss of generality that $y$ is preferred to $y'$ given $x$ according to the reward model, that is $R(x,y)>R(x,y')$, and redefine $R(x,y) = - R(x,y') = \frac{1}{4}$, then we have 
    \begin{align}
        \nabla \ell_\copg (x,y,y';\pi) = -\frac{1}{2\beta} \left(\frac{1}{2} - \beta \left(\ln\frac{\pi(y|x)}{\piref(y|x)} - \ln\frac{\pi(y'|x)}{\piref(y'|x)}\right)\right)^2,
    \end{align}
    where the term on the right-hand side is the gradient of the sample-based IPO loss to be minimized.
\end{prop}
\begin{proof}
    First, from Eq.~\eqref{eq:proof:grad_copg}, we have that
    \begin{equation}
        \nabla \ell_\copg(x,y,y';\pi) = \left(R_{\beta}^\pi(x,y) - R_{\beta}^\pi(x,y')\right) \nabla \ln \pi(y|x)
        + \left(R_{\beta}^\pi(x,y') - R_{\beta}^\pi(x,y)\right) \nabla \ln \pi(y'|x).
    \end{equation}
    Given the assumptions ($y$ preferred to $y'$ given $x$ and binarized reward, that is redefine $R(x,y) = - R(x,y') = -\frac{1}{4}$), and given the definition of $R_\beta^\pi$ in Eq.~\eqref{eq:reg_reward_full}, the gradient writes
    \begin{equation}
        \nabla \ell_\copg(x,y,y';\pi) = \left(\frac{1}{2} - \beta \ln\frac{\pi(y|x)}{\piref(y|x)} + \beta \ln\frac{\pi(y'|x)}{\piref(y'|x)}\right) \left(\nabla\ln \pi(y|x) - \nabla\ln \pi(y'|x)\right).
    \end{equation}
    Now, let consider the gradient of the sample-based IPO loss:
    \begin{align}
        &\nabla \left(\frac{1}{2} - \beta \left(\ln\frac{\pi(y|x)}{\piref(y|x)} - \ln\frac{\pi(y'|x)}{\piref(y'|x)}\right)\right)^2
        \\
        = &2 \left(\frac{1}{2} - \beta \left(\ln\frac{\pi(y|x)}{\piref(y|x)} - \ln\frac{\pi(y'|x)}{\piref(y'|x)}\right)\right)\left(\beta \nabla \ln \pi(y'|x) - \beta \nabla \ln \pi(y|x)\right)
        \\
        = &- 2 \beta \nabla \ell_\copg(x,y,y';\pi).
    \end{align}
    This proves the stated result.
\end{proof}

\newpage

\section{Qualitative results}
\label{appx:gen_examples}

We list two examples from the TL;DR dataset during evaluation. We observe that the summary improves in quality during training, with more elaborate structures. Additionally, we started noticing some reward hacking artifacts after 2 epochs ($2\times 600$ steps) of the training.

\begin{tcolorbox}
\underline{SUBREDDIT}: relationships\\
\underline{TITLE}: (24/F) I hate losing arguments to my bf (27/m)\\
\underline{POST}: I just got into another heated petty argument with my bf of 3 years! He started telling our roommate a story about someone, I was sure he was thinking of something else, he stops dead in his tracks and tells me that that's not what happened. I said more details of what I remembered, he told me again that I was wrong and I should believe him because it's a story that happened to him. He said what he always does, "why is it that when we disagree you assume I must be wrong? Why don't you ever consider that you could be wrong?" 

By that point I had realized it was possible I was wrong, but I absolutely did not want to concede that to him or offer him any slack. I don't get like this with everybody, but my bf and I both have this stupid trait of never backing down! And I hate admitting I'm wrong to him because I feel like he acts super smug and condescending, when I make a mistake he laughs at me in an "oh, you" sort of way that INFURIATES me. I know I overreact to this sort of thing but I don't know how to improve!
\end{tcolorbox}

\textbf{DPO:}
\begin{itemize}
    \item \emph{Step=300:} I hate admitting to my bf that I’m wrong and I get super defensive when he acts like he’s so happy and smug that I was wrong. How do I improve?
    \item \emph{Step=600:} I hate admitting I’m wrong to my bf because he acts super smug and I don’t like it. How do I get better about this?
    \item \emph{Step=900:} I hate admitting I’m wrong to my bf in arguments because he acts super smug and I can’t stand it, but I also hate losing arguments so I always fight to the death with him even when I’m sure I’m wrong sometimes! What do I do?!
    \item \emph{Step=1200:} I hate admitting I’m wrong to my bf because he acts super smug and I want to win every argument. How do I improve ?!?!?!?!?!?!?!?!?!?!?!?!?!?!?!?!?!?!?!?!?!?!?!?!?!?!?!?\break!?!?!?!?!?!?!?!?!?!?!?!?!?!?!?!?!?!?!?!?!!?!?!?!?!?!?!?!?!?!?!?!?!?!?!?!?!?!?!?!?!?!?!?!?!?!?!?!!?!?!?\break!?!?!?!?!?!?!?!?!?!?!?!?!?!?!?!?!?!?!?!!?!?!?!?!!?!?!?!?!?!?!?!?!?!?!?!?!?!?!?!?!?!?!?!?!?!?!?!?!?!?!?\break!?!?!?!?!?!?!?!?!?!?!?!?!?!?!?!?!?!?!?!?!
\end{itemize}   

\textbf{IPO:}
\begin{itemize}
    \item \emph{Step=300:} I hate admitting to my bf that I’m wrong and I get super defensive when he acts like he’s so happy that he was right and I was wrong. I want to improve so I don’t get so upset over this!
    \item \emph{Step=600:} I hate admitting I’m wrong to my bf because he acts super smug and I don’t like it, but I also hate losing arguments and always want to be right. I need to learn how to not care so much!
    \item \emph{Step=900:} I hate admitting I’m wrong to my bf in arguments because he acts super smug and I can’t stand it, but I also hate losing arguments! How do I improve so I don’t feel so defensive about being wrong?
    \item \emph{Step=1200:} I hate admitting I’m wrong to my bf in arguments because he acts super smug and I want to win every argument anyway even if I’m not 100\% sure I’m right. How do I stop?
\end{itemize}   

\textbf{\copg:}
\begin{itemize}
    \item \emph{Step=300:} I hate admitting I'm wrong to my bf in arguments, he always acts super smug and I can't stand it. How can I get better about this?
    \item \emph{Step=600:} I hate admitting I'm wrong to my bf and I hate when he acts super smug and condescending when he knows he has proven he was right and I was wrong. I need to learn how to not let this get to me!
    \item \emph{Step=900:} I hate admitting defeat in petty arguments to my bf, he always acts so smug and I can't stand it, but I don't know how to improve so I don't get as upset with him over it!
    \item \emph{Step=1200:}  I hate admitting that I'm wrong to my bf after a disagreement because he acts like a superior being, I don't want to back down and admit it to him, but I also don't want to continue the argument because it just gets more heated and nasty.
\end{itemize}

\vspace{3em}
\begin{tcolorbox}
\underline{SUBREDDIT:} relationships \\
\underline{TITLE:} Me [19 F] with my boyfriend [19 M ] 3.5 years, when is it time to let go? with line breaks. \\
\underline{POST:} Okay, so I've never really posted anything before, so sorry if this is weird or I'm not doing it right or something. I started dating my current BF when I was 15 and a sophomore in high school, he was 16 and a junior. Now we are both in college. We are insanely close. Like spend literally ALL our time together. He is incredibly important to me and we do everything together. I love him a lot and I have incredible respect for him as a person. 
The problem is we both don't have friends outside of the relationship. 
I mean we both have friends. We have a lot of friends really, but we just don't feel like we connect with anyone other than each other anymore. It's hard to explain. 
We both ditched all of our friends from high school. Some of it was because of our relationship, but a lot of it was high school is shitty. 
\end{tcolorbox}

\textbf{DPO:}
\begin{itemize}
    \item \emph{Step=300:} I know I need to break up with my boyfriend of 3.5 years, but I’m scared because I feel like I won’t ever connect with anyone again because we have kind of abandoned our friends for each other.
    \item \emph{Step=600:} I know I need to break up with my boyfriend of 3.5 years, but I feel like I won’t ever connect with anyone else because I’ve been so focused on this relationship my whole life. I don’t know what to do.
    \item \emph{Step=900:} I know I need to break up with my boyfriend I’ve been with for 3.5 years, but I’m afraid because I feel like I won’t ever connect with anyone again after him. I feel like I only connect with him. I don’t feel like I connect with anyone else.
    \item \emph{Step=1200:} My boyfriend and I have pretty much zero outside friends and we are too dependent on each other, I know we should break up but I’m scared I will never connect with anyone else. I’d really appreciate some advice. Thanks.
\end{itemize}

\textbf{IPO:}
\begin{itemize}
    \item \emph{Step=300:} I know I need to break up with my boyfriend of 3.5 years, but I’m scared because I feel like I won’t ever connect with anyone outside of my SO now. Has anyone expressed something like this?
    \item \emph{Step=600:} I know I need to end my relationship with my boyfriend whom I love very dearly, but I’m scared because I feel like I won’t ever connect with anyone outside of the relationship anymore if I do.
    \item \emph{Step=900:} I know it’s time to move on from my long-term relationship, but I’m scared because I don’t feel like I’ll ever connect with anyone again outside of my SO. Has anyone else expressed this feeling before? If so, what did you do?
    \item \emph{Step=1200:} My boyfriend and I have pretty much ditched all of our friends from high school and replaced them with each other. We know we need to break up eventually, but I’m scared because I don’t feel like I’ll ever connect with anyone else. Has anyone else experienced something like this?
\end{itemize}   

\textbf{\copg:}
\begin{itemize}
    \item \emph{Step=300:} I know I need to break up with my boyfriend of 3.5 years whom I love very dearly, but I'm scared because I feel like I won't ever connect with anyone else outside of this relationship. Has anyone expressed this feeling before?
    \item \emph{Step=600:} I know I need to end my relationship with my boyfriend whom I love very dearly, but I am scared I'll never find anyone I connect with after him because I feel like I don't connect with anyone other than him.
    \item \emph{Step=900:} I know I need to end my long-term relationship, but I'm scared because I don't feel like I'll ever be able to connect with anyone else because all of my friends currently are either from this relationship or I made this year and don't feel like I connect with them.
    \item \emph{Step=1200:} My boyfriend and I are pretty much all each other have ever cared about or focused on in terms of friends. I know I need to break up with him, but I'm scared I won't find any friends I connect with ever again.
\end{itemize}

\end{document}